\documentclass[runningheads]{llncs}
\usepackage{graphicx}
\usepackage[colorlinks=true, allcolors=blue]{hyperref}
\usepackage{amsmath,amsfonts,amssymb}
\usepackage{algorithm2e} 
\usepackage{algpseudocode} 
\usepackage[T1]{fontenc} 
\SetKwInput{KwInput}{Input}                
\SetKwInput{KwOutput}{Output}              
\usepackage[normalem]{ulem}

\newcommand{\MKeditConfirmed}[2]{{\color{black}#2}}
\newcommand{\R}{\mathbb{R}}
\newcommand{\V}{\mathcal{V}}

\newcommand{\ignore}[1]{}
\newcommand{\samethanks}[1][\value{footnote}]{\footnotemark[#1]}
\usepackage{amsmath}
\DeclareMathOperator*{\argmax}{arg\,max}

\begin{document}

\title{ Skin Lesion Correspondence Localization in Total Body Photography }

\author{Wei-Lun Huang\inst{1}\thanks{Corresponding author: Wei-Lun Huang (whuang44@jh.edu)} \and
Davood Tashayyod\inst{5} \and
Jun Kang\inst{3}\and
Amir Gandjbakhche\inst{4} \and
Michael Kazhdan\inst{1}\thanks{Co-senior authors} \and
Mehran Armand\inst{1,2}\samethanks[2]
}
\authorrunning{W. Huang et al.}
\institute{Johns Hopkins University, Department of Computer Science, Baltimore, MD, USA \and
Johns Hopkins University, Department of  Orthopaedic Surgery, Baltimore, MD, USA \and
Johns Hopkins School of Medicine, Department of Dermatology, Baltimore, MD, USA \and
Eunice Kennedy Shriver National Institute of Child Health and Human Development, Bethesda, MD, USA \and
Lumo Imaging, Rockville, MD, USA
}

\maketitle              
\begin{abstract}
Longitudinal tracking of skin lesions - finding correspondence, changes in morphology, and texture - is beneficial to the early detection of melanoma. However, it has not been well investigated in the context of full-body imaging. We propose\MKeditConfirmed{d}{} a novel framework combining geometric and texture information to localize skin lesion correspondence from a source scan to a target scan in total body photography (TBP). Body landmarks or sparse correspondence are first created on the source and target 3D textured meshes. Every vertex on each of the meshes is then mapped to a feature vector characterizing the geodesic distances to the landmarks on that mesh. Then, for each lesion of interest (LOI) on the source, its corresponding location on the target is first coarsely estimated using the geometric information encoded in the feature vectors and then refined using the texture information. We evaluated the framework quantitatively on both a public and a private dataset, for which our success rates (at 10 mm criterion) are comparable to the only reported longitudinal study. As full-body 3D capture becomes more prevalent and has higher quality, we expect the proposed method to constitute a valuable step in the longitudinal tracking of skin lesions.

\keywords{Total body photography  \and Skin lesion longitudinal tracking \and 3D correspondence.}
\end{abstract}

\section{Introduction}
\label{sec:intro}
Evolution, the change of pigmented skin lesions, is a risk factor for melanoma \cite{abbasi2004early}. Therefore, longitudinal tracking of skin lesions over the whole body is beneficial for early detection of melanoma \cite{halpern2003total}. However, establishing skin lesion correspondences across multiple scans from different patient visits has not been well investigated in the context of full-body imaging.

Several techniques have been proposed to match skin lesions across pairs of 2D images \cite{li2016skin,mcgregor1998automatic,mirzaalian2009graph,mirzaalian2012uncertainty,mirzaalian2013spatial,mirzaalian2016skin,perednia1992automatic,roning1998registration,white1992automatic}. Early work used geometric constraint\MKeditConfirmed{}{s} imposed by initial matches of skin lesions (manual selection or automatic detection) to align images and further match other skin lesions \cite{mcgregor1998automatic,perednia1992automatic,roning1998registration,white1992automatic}. Mirzaalian and colleagues published a series of works for establishing lesion correspondence in image space \cite{mirzaalian2009graph,mirzaalian2012uncertainty,mirzaalian2013spatial,mirzaalian2016skin}. Li et al. \cite{li2016skin} used a CNN to output a 2D vector field for pixel-wise correspondences between the two input images. Though effective at matching skin lesions across pairs of images, the extension of these methods to the context of total body photography (TBP) for longitudinal tracking remains a challenge.

Several works have been proposed for tackling the skin lesion tracking problem over the full body \cite{korotkov2018improved,korotkov2014new,strkakowska2022skin,strzelecki2021skin}. However, they are either only applicable in well-controlled environments or do not extend to the tracking of lesions across scans at different visits. Recently, the concept of finding lesion correspondence using a 3D representation of the human body has been explored in \cite{zhao2022skin3d} and \cite{bogo2014automated} by using a template mesh. However, accurately deforming a template mesh to fit varying body shapes is challenging when the scanned shape deviates from the template, leading to large errors in downstream tasks such as establishing shape correspondence. Additionally, \cite{zhao2022skin3d} does not take advantage of texture, while \cite{bogo2014automated} uses texture in a common UV map that may lead to failures when geodesically close locations on the surface are mapped to distant sites in the texture map (e.g. when the two locations are on opposite sides of a texture seam).

\begin{figure} [ht]
   \begin{center}
   \begin{tabular}{c} 
    \includegraphics[width=\textwidth]{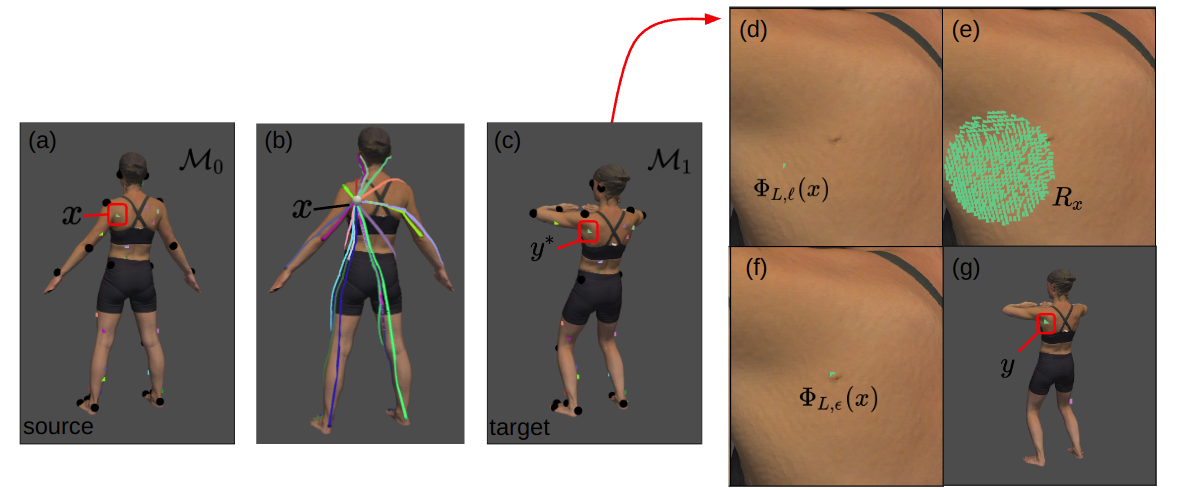}
    \end{tabular}
    \end{center}
    \caption{Visualization of the correspondence localization workflow with geometric and texture information. (a) and (c) show a source and a target textured mesh with landmarks (black dots) and labeled lesions of interest (color dots). (b) shows the geodesic paths from a lesion $x$ in (a) to all the landmarks. (d) shows the correspondence of the lesion $\Phi_{L,\ell}(x)$ derived from the geometric feature descriptors. (e) shows the local region $R_x$ for refining the position of the correspondence. (f) shows the correspondence of the lesion $\Phi_{L,\epsilon}(x)$ from the local texture descriptors. (g) shows the lesion correspondences with confidence and anchored as new landmarks after one iteration.}
    \label{fig:localization workflow}
\end{figure}

We propose a novel framework for finding skin lesion correspondence iteratively using geometric and texture information (Fig.~\ref{fig:localization workflow}). We demonstrate the effectiveness of the proposed method in localizing lesion correspondence across scans in a manner that is robust to changes in body pose and camera viewing directions. Our code is available at \url{https://github.com/weilunhuang-jhu/LesionCorrespondenceTBP3D}.

\section{Methods}
Given a set of lesions of interest (LOIs) $X$ in the source mesh, we would like to find their corresponding positions $Y$ in the target mesh. Formally, we assume we are given source and target meshes, $\mathcal{M}_0$ and $\mathcal{M}_1$, with vertex sets $\mathcal{V}_k\subset\mathcal{M}_k$ and corresponding landmark sets $L_k\subset\mathcal{V}_k$ with $|L_0|=|L_1|=S$.
We achieve this by computing a dense correspondence map $\Phi_{L,\epsilon}:\V_0\rightarrow\V_1$, initially defined using geometric information and refined using textural information. Then we use that to define a map taking lesions of interest on the source to positions on the target $\Phi:X\rightarrow\V_1$.

\subsection{Landmark-based correspondences}
\label{subsection:shape feature}
We define an initial dense correspondence between source and target vertices by leveraging the sparse landmark correspondences \cite{datar2013geodesic,kim2021deep}. We do this by mapping source and target vertices into a high-dimensional space, based on their proximity to the landmarks, and then comparing positions in the high-dimensional space.

Concretely, we define maps $\ell_k:\mathcal{V}_k\rightarrow\R^S$, associating a vertex $v\in\V_k$ with an $S$-dimensional feature descriptor that describes the position of $v$ relative to the landmarks:
\begin{equation}\label{eq:shape feature}
\ell_k(v) = \left[\frac1{D_k(v,l_1)},\cdots,\frac1{D_k(v,l_S)}\right]^\top
\end{equation}
where $D_k:\mathcal{M}_k\times\mathcal{M}_k\rightarrow\R^{\geq0}$ is the geodesic distance function on $\mathcal{M}_k$. We use the reciprocal of geodesic distance so that landmarks closer to $v$ contribute more significantly to the feature vector.

Given this mapping, we create an initial dense correspondence between the source and target vertices, $\Phi_{L,\ell}:\mathcal{V}_0\rightarrow\mathcal{V}_1$ by mapping a source vertex $v\in \V_0$ to the target vertex with the most similar feature descriptor (with similarity measured in terms of the normalized cross-correlation):
\begin{equation}\label{eqn:correspondence from shape}
\Phi_{L,\ell}(v) = \argmax_{v' \in \V_1} \left\{C_{L,\ell}(v,v') = \frac{\langle \ell_0(v),\ell_1(v')\rangle}{||\ell_0(v)||\cdot||\ell_1(v')||}\right\}.  
\end{equation}

\subsection{Texture-based refinement}
\label{subsection:local texture}

While feature descriptors of corresponding vertices on the source and target mesh are identical when 1) the landmarks are in perfect correspondence, and 2) the source and target differ by an isometry, neither of these assumptions holds in real-world data. To address this, we use local texture to assign an additional feature descriptor to each vertex and use these texture-based descriptors to refine the coarse correspondence given by $\Phi_{L,\ell}:\V_0\rightarrow\V_1$. Various texture descriptors have been proposed, e.g. SHOT \cite{tombari2010unique,tombari2011combined,salti2014shot}, RoPS\cite{guo2013rotational}, and ECHO\cite{mitchel2021echo}. We selected the ECHO descriptor for its better descriptiveness and robustness to noise.

Letting $\epsilon_k(v)\in\R^N$ denote the ECHO descriptor of vertex $v\in\V_k$, our goal is to refine the dense correspondence so that corresponding source and target vertices also have similar descriptors. However, to avoid problems with repeating (local) textures, we would also like the correspondence to stay close to the correspondence defined by the landmarks.

We achieve this as follows: To every source vertex $v\in\V_0$ we associate a region $R_v\subset\V_1$ of target vertices that are either close to $\Phi_{L,\ell}(v)$ (the corresponding vertex on $\V_1$ as predicted by the landmarks) or have similar geometric feature descriptors:
\begin{equation}\label{eqn:search region}
R_v=
\left\{v'\in\V_1\Big|\,D_1\left(v',\Phi_{L,\ell}(v)\right)<\varepsilon_1\hbox{ or }
C_{L,\ell}(v', \Phi_{L,\ell}(v))>\varepsilon_2\right\}.   
\end{equation}

Given this region, we define the target vertex corresponding to a source as the vertex within the region that has the most similar ECHO descriptor (using the normalized cross-correlation as before).

In practice, we compute the ECHO descriptor over three different radii, obtaining three descriptors for each vertex, $\epsilon_k^1(v),\epsilon_k^2(v),\epsilon_k^3(v)\in\R^N$. The selection of three different radii in ECHO descriptors is done to accommodate different sizes of lesions and their surrounding texture, and the values are empirically determined. This gives a mapping $\Phi_{L,\epsilon}:\V_0\rightarrow\V_1$ defined in terms of the weighted sum of cross-correlations:
\begin{equation}\label{eqn:correspondence from texture}
\Phi_{L,\epsilon}(v)=\argmax_{v'\in R_v}\left\{C_{L,\epsilon}(v, v') =  \sum_{i=1}^3 w_i\frac{\langle\epsilon_0^i(v),\epsilon_1^i(v')\rangle}{\|\epsilon_0^i(v)\|\cdot\|\epsilon_1^i(v')\|}\right\} \ ,
\end{equation}
where $C_{L,\epsilon}(v,v') \in [0,1]$ is the texture score of the target vertex $v'$ and $w_i$ is the weight of the cross-correlation between the ECHO descriptors computed at each radius.

\subsection{Iterative Skin Lesion Correspondence Localization Framework}
\label{subsection:iterative framework}

\begin{figure} [ht]
   \begin{center}
   \begin{tabular}{c} 
   \includegraphics[width=\textwidth]{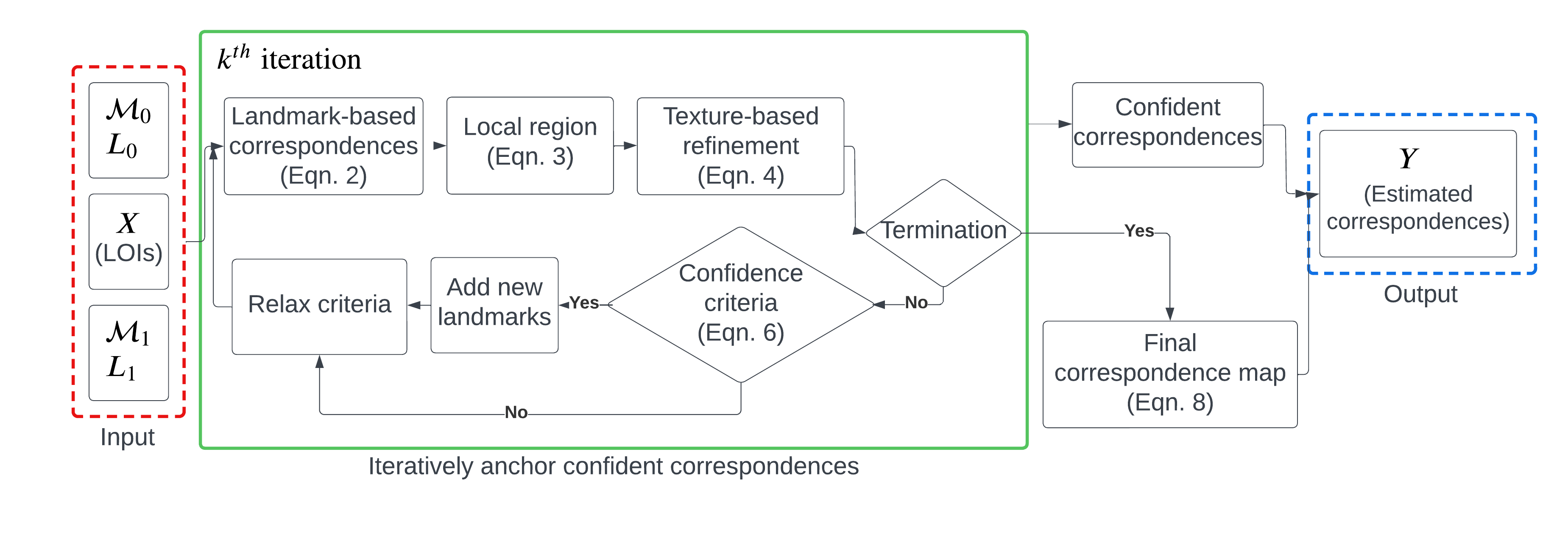}
	\end{tabular}
	\end{center}
   \caption{Block diagram of the iterative lesion correspondence localization algorithm.}
   \label{fig:block diagram}
\end{figure}

While each source LOI has a corresponding position on the target mesh as given by $\Phi_{L,\epsilon}:\V_0\rightarrow\V_1$, not all correspondences are localized with high confidence when 1) the local texture is not well-preserved across scans and 2) the local region $R_v$ does not include the true correspondence.
To address this, we adapt our algorithm for computing the correspondence map $\Phi:X\rightarrow\V_1$ by iteratively growing the set of landmarks to include LOI correspondences about which we are confident, similar to the way in which a human annotator would label lesion correspondence (Fig.~\ref{fig:block diagram}).

\subsubsection*{Iteratively anchor confident correspondences}

We iteratively compute correspondence maps $\Phi^k_{L,\ell}:\V_0\rightarrow\V_1$ and $\Phi^k_{L,\epsilon}:\V_0\rightarrow\V_1$, with the superscript denoting the $k^{th}$ iteration. For each map $\Phi^k_{L,\epsilon}$ and every LOI $x\in X$, we determine if we are confident in the correspondence $\{x,\Phi^k_{L,\epsilon}(x)\}$ by evaluating a binary function $\chi^k_L:X\rightarrow\{0,1\}$. Denoting by $X'$ the subset of LOIs about which we are confident, we add the pairs $\{x',\Phi^k_{L,\epsilon}(x')\}$ to the landmark set $L$ and remove the LOI $x'\in X'$ from $X$. We iterate this process until all the correspondences of LOIs are confidently found or a maximum number of iterations ($K$) have been performed. 

Lesion correspondence confidence is measured using three criteria: i) texture similarity, ii) agreement between geometric and textural correspondences, and iii) the unique existence of a similar lesion within a region. To quantify uniqueness, we compute the set of target vertices whose textural descriptor is similar to that of the LOI:
\begin{equation}\label{eq:high texture similarity set}
S_x^\delta=\left\{v'\in R_x\big|\, C_{L,\epsilon}(x,v')>\delta\right\} \ ,  
\end{equation}
and consider the diameter of the set (defined in terms of the mean of the distances of vertices in $S_x^\delta$ from the centroid of $S_x^\delta$).
Putting this together, we define confidence as
\begin{equation}\label{eqn:criteria}
\chi^k_L(x) =
C_{L,\epsilon}\big(x,\Phi^k_{L,\epsilon}(x)\big) > \varepsilon_3
\quad\vee\quad
D_1\big(\Phi^k_{L,\ell}(x),\Phi^k_{L,\epsilon}(x)\big)<\varepsilon_4
\quad\vee\quad
\varnothing(S_x^\delta)<\varepsilon_5
,
\end{equation}
where the initial values of thresholds $\varepsilon_i$ are empirically chosen. To further support establishing correspondences, we relax the thresholds $\varepsilon_i$ in subsequent iterations, allowing us to consider correspondences that are further away and about which we are less confident.

\subsubsection*{Final correspondence map} Having mapped every high-confidence LOI to a corresponding target vertex, we must complete the correspondence for the remaining low-confidence LOIs. We note that for a low-confidence LOI $x\in X$, the texture in the source mesh is not well-matched to the texture in the target, for any $v'\in R_x$. (Otherwise the first term in $\chi_L^k$ would be large.) 

To address this, we would like to focus on landmark-based similarity. However, by definition of $R_x$, for all $v'\in R_x$, we know that the landmark descriptors of $x$ and $v'$ will all be similar, so that $C_{L,\ell}$ will not be discriminating. Instead, we use a standard transformation to turn distances into similarities. Specifically, we define geometric score between a source LOI $x$ and a target vertex $v'\in R_x$ in terms of the geodesic distance between $v'$ and the corresponding position of $x$ in $\V_1$, as predicted by the landmark descriptors:
\begin{equation} \label{eq:shape score}
C_L(x, v') = e^{-\frac{1}{2 \sigma^{2}}D_1^2\left(v',\ \Phi^{K}_{L,\ell}(x)\right)} \in [0,1] \ ,
\end{equation}
where $\sigma$ is the maximum geodesic distance from a vertex within $R_x$ to $\Phi^{K}_{L,\ell}(x)$.

Therefore, for a remaining LOI, we define its corresponding target vertex as the vertex with the highest weighted sum of the geometric and texture scores:
\begin{equation}\label{eqn:correspondence from combined}
\Phi(x) = \argmax_{v'\in R_x}\big\{ w_1\cdot C_L(x, v') + w_2\cdot C_{L,\varepsilon}(x, v')\big\}\ ,
\end{equation}
where $w_1$ and $w_2$ are the weights for combining the scores.

\section{Evaluation and Discussion}
\label{sec:evaluation}

\subsection{Dataset}
\label{sec:dataset}
We evaluated our methods on two datasets. The first dataset is from Skin3D \cite{zhao2022skin3d} (annotated 3DBodyTex \cite{saint20183dbodytex,saint2019bodyfitr}). The second dataset comes from a 2D Imaging-Rich Total Body Photography system (IRTBP), from which the 3D textured meshes are derived from photogrammetry 3D reconstruction. The number of vertices is on average 300K and 600K for Skin3D and IRTBP datasets respectively. The runtime using 10 iterations is several minutes (on average) on an Intel i7-11857G7 processor. Example data of the two datasets can be found in the supplement. 

\subsection{Correspondence Localization Error and Success Rate}
\label{sec:cle and success rate}
\begin{figure} [ht]
   \begin{center}
   \begin{tabular}{c} 
   \includegraphics[width=\textwidth]{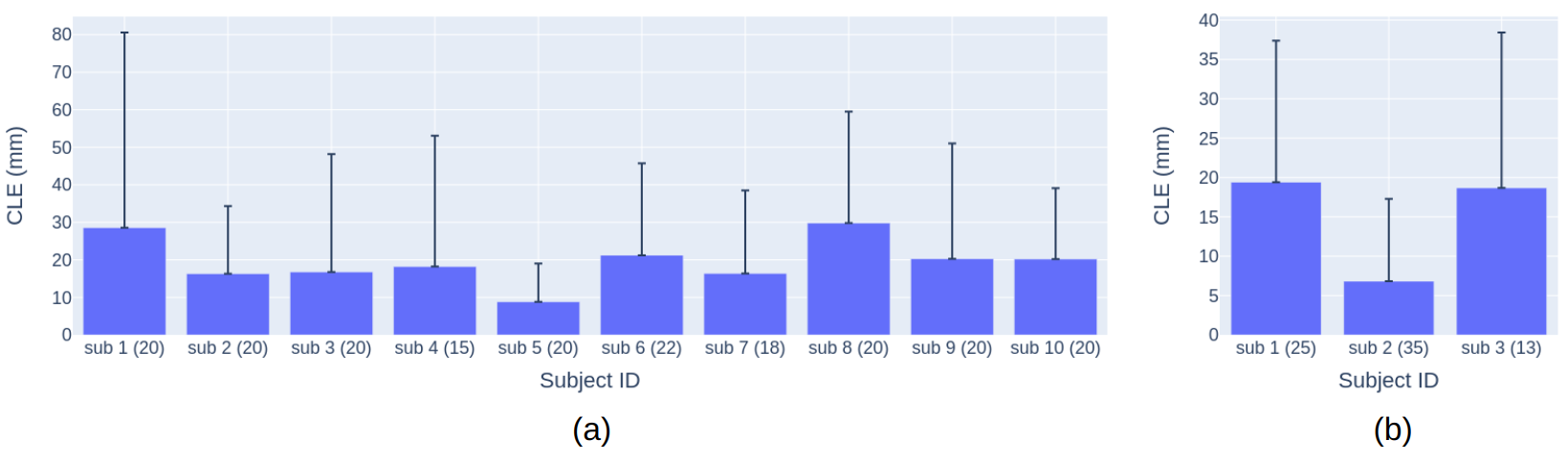}
	\end{tabular}
	\end{center}
   \caption[example] 
   { \label{fig:cle}
Mean and standard deviation of correspondence localization error (CLE) for individual subjects in (a) Skin3D and (b) IRTBP datasets evaluated with the proposed methods. There are 10 and 3 subjects in the Skin3D and IRTBP datasets respectively. The number of LOIs for individual subjects is included in the parentheses.}
\end{figure}

Average correspondence localization error (CLE) for individual subjects, defined as the geodesic distance between the ground-truth and the estimated lesion correspondence, is shown in Fig.~\ref{fig:cle}. To interpret CLE in a clinical application, the localized correspondence is successful if its CLE is less than a threshold criterion. We measured the success rate as the percentage of the correctly localized skin lesions over the total number of skin lesion pairs in the dataset.

To compare our result to the existing method \cite{zhao2022skin3d}, we compute our success rates with the threshold criterion at 10 mm. As shown in table~\ref{tab:accuracy comparison}, the performance of our method is comparable to the previously reported longitudinal accuracy. The qualitative result of the localized correspondence in the Skin3D dataset is shown in Fig.~\ref{fig:qualitative_result}. A table of parameters used in the experiments can be found in the supplement. 

\begin{figure} [ht]
  \begin{center}
  \begin{tabular}{c} 
  \includegraphics[width=11.5cm] {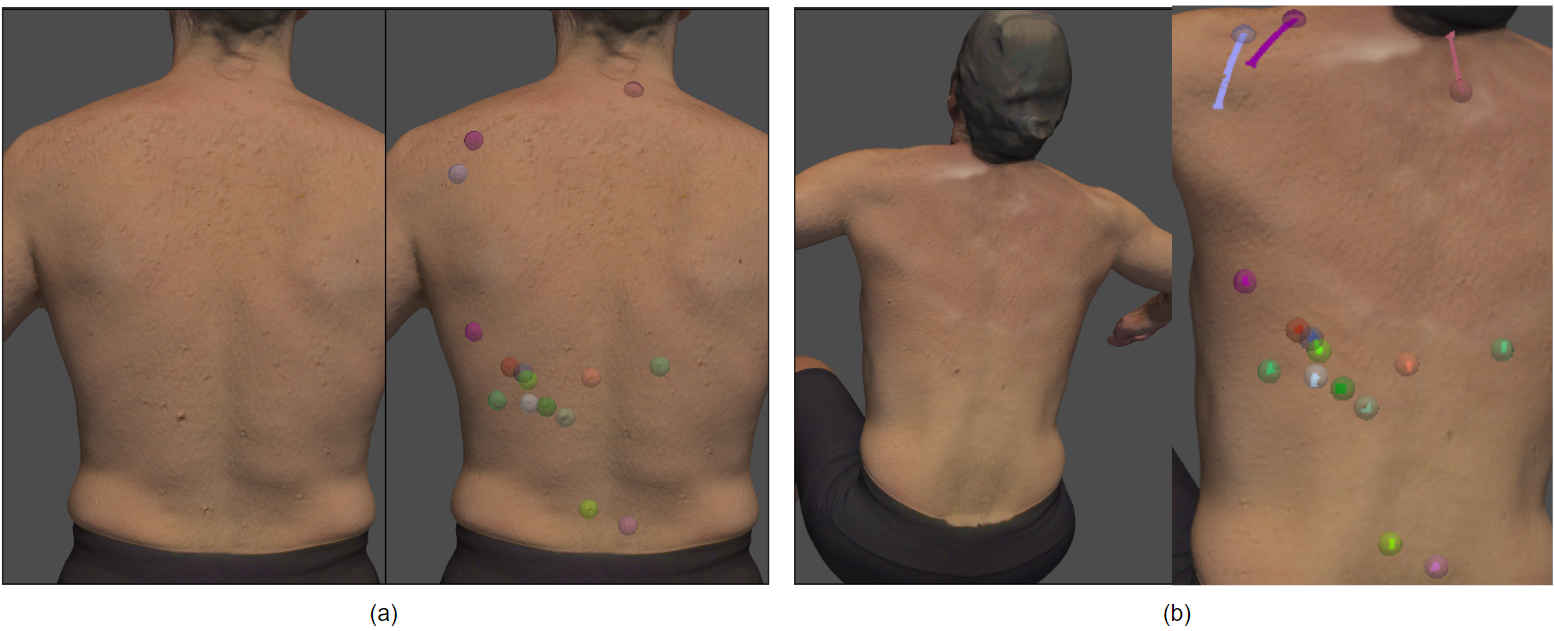}
	\end{tabular}
	\end{center}
  \caption[example] 
  { \label{fig:qualitative_result}
The qualitative result on Skin3D dataset. (a) shows a source scan (left) and the scan with annotated LOIs (right). (b) shows a target scan (left) and the scan with annotated LOIs (transparent sphere) and the estimated correspondence of LOIs (solid dot) (right). The lesion correspondence pairs are shown in the same color. Large CLE occurs when the local texture is not well-preserved across scans.}
\end{figure}

\begin{table}[ht]
\caption{Comparison of the success rate on Skin3D dataset. Each metric is computed on a pair of meshes (for one subject) and averaged across paired meshes with the standard deviation shown in brackets. The method \textbf{\textit{Texture radius 50}} and \textbf{\textit{Combined radius 50}} are defined in Fig.~\ref{fig:skin3d_result}.} 
\label{tab:accuracy comparison}
\resizebox{\textwidth}{!}{    
\begin{tabular}{|l|l|l|l|l|} 
\hline
\rule[-1ex]{0pt}{3.5ex}   & Skin3D \cite{zhao2022skin3d}
& Texture radius 50 & Combined radius 50 & Iterative algorithm \\
\hline
\rule[-1ex]{0pt}{3.5ex}  Success rate  & 0.5 (0.38) & 0.48 (0.17) & 0.45 (0.14) & 0.57 (0.14)\\
\hline
\end{tabular}
}
\end{table}

\subsection{Usage of Texture on 3D Surface} \label{sec:usage of texture}
\begin{figure}[ht]
   \begin{center}
   \begin{tabular}{c} 
   \includegraphics[width=\textwidth]{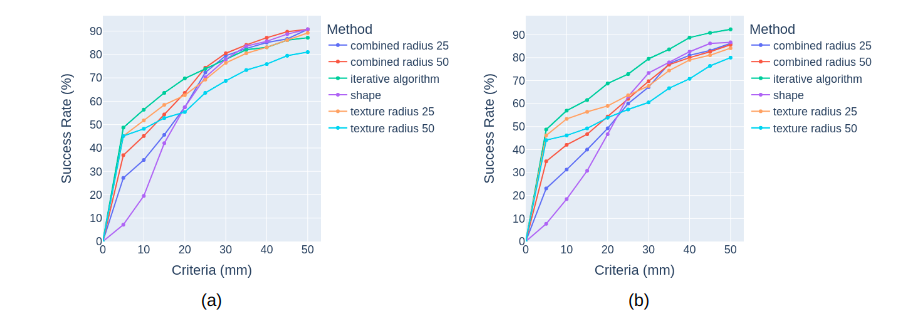}
	\end{tabular}
	\end{center}
   \caption[example] 
   { \label{fig:skin3d_result}
   Success rate under different criteria for the proposed algorithm on the Skin3D dataset. (a) and (b) show the results of two combinations of source and target meshes. \textbf{\textit{Iterative algorithm}} is the proposed algorithm with the anchor mechanism. \textbf{\textit{Shape}} is the method using $\Phi^1_{L,\ell}$. \textbf{\textit{Texture radius 25}} and \textbf{\textit{texture radius 50}} are the methods using $\Phi^1_{L,\epsilon}$ with $\varepsilon_1$ (Eqn.~\ref{eqn:search region}) selected at 25 mm and 50 mm. \textbf{\textit{Combined radius 25}} and \textbf{\textit{combined radius 50}} are the methods using Eqn.~\ref{eqn:correspondence from combined} with one iteration.
   }
\end{figure}

We believe that the geometric descriptor only provides a coarse correspondence while the local texture is more discriminating. Fig.~\ref{fig:skin3d_result} shows the success rate under different threshold criteria for the proposed methods. Since we have two combinations of defining source and target for two scans, we measured the result in both to ensure consistency (Fig.~\ref{fig:skin3d_result} (a) and (b)). As expected, we observed that using geometric information with body landmarks and geodesic distances is insufficient for localizing lesion correspondence accurately. However, the correspondence map $\Phi_{L,\epsilon}$ refined with local texture may lead to correspondences with large CLE when using a large region $R_x$. The figure shows the discriminating power and the large-error-prone property of using $\Phi_{L,\epsilon}$ to localize lesion correspondence with one iteration (relatively high success rates under strict criteria and relatively low success rates under loose criteria, compared to the correspondence map combining geometric and texture scores in Eqn.~\ref{eqn:correspondence from combined}). The figure also shows the effectiveness of the proposed algorithm when the iterative anchor mechanism is used to localize lesion correspondence, having consistently higher success rates with the criteria within 20 mm. 

\section{Conclusions and Limitations}
The evolution of a skin lesion is an important sign of a potentially cancerous growth and total body photography is useful to keep track of skin lesions longitudinally. We proposed a novel framework that leverages geometric and texture information to effectively find lesion correspondence across TBP scans. The framework is evaluated on a private dataset and a public dataset with success rates that are comparable to those of the state-of-the-art method.

The proposed method assumes that the local texture enclosing the lesion and its surroundings should be similar from scan to scan. This may not hold when the appearance of the lesion changes dramatically (e.g. if the person acquires a tattoo). Also, the resolution of the mesh affects the precision of the positions of landmarks and lesions. In addition, the method may not work well with longitudinal data that has non-isometric deformation due to huge variations in body shape, inconsistent 3D reconstruction, or a dramatic change in pose and, therefore, topology, such as an open armpit versus a closed one.

In the future, the method needs to be evaluated on longitudinal data with longer duration and new lesions absent in the target. In addition, an automatic method to determine accurate landmarks is desirable. Note that although we rely on the manual selection of landmarks, the framework is still preferable over manually annotating lesion correspondences when a subject has hundreds of lesions. As the 3D capture of the full body becomes more prevalent with better quality in TBP, we expect that the proposed method will serve as a valuable step for the longitudinal tracking of skin lesions.

\subsubsection{Acknowledgments}
The research was in part supported by the Intramural Research Program (IRP) of the NIH/NICHD, Phase I of NSF STTR grant 2127051, and Phase I NIH/NIBIB STTR grant R41EB032304.
\bibliographystyle{splncs04}
\bibliography{references}

\begin{thebibliography}{10}
\providecommand{\url}[1]{\texttt{#1}}
\providecommand{\urlprefix}{URL }
\providecommand{\doi}[1]{https://doi.org/#1}

\bibitem{abbasi2004early}
Abbasi, N.R., Shaw, H.M., Rigel, D.S., Friedman, R.J., McCarthy, W.H., Osman,
  I., Kopf, A.W., Polsky, D.: Early diagnosis of cutaneous melanoma: revisiting
  the abcd criteria. Jama  \textbf{292}(22),  2771--2776 (2004)

\bibitem{bogo2014automated}
Bogo, F., Romero, J., Peserico, E., Black, M.J.: Automated detection of new or
  evolving melanocytic lesions using a 3d body model. In: International
  Conference on Medical Image Computing and Computer-Assisted Intervention. pp.
  593--600. Springer (2014)

\bibitem{datar2013geodesic}
Datar, M., Lyu, I., Kim, S., Cates, J., Styner, M.A., Whitaker, R.: Geodesic
  distances to landmarks for dense correspondence on ensembles of complex
  shapes. In: Medical Image Computing and Computer-Assisted
  Intervention--MICCAI 2013: 16th International Conference, Nagoya, Japan,
  September 22-26, 2013, Proceedings, Part II 16. pp. 19--26. Springer (2013)

\bibitem{guo2013rotational}
Guo, Y., Sohel, F., Bennamoun, M., Lu, M., Wan, J.: Rotational projection
  statistics for 3d local surface description and object recognition.
  International journal of computer vision  \textbf{105}(1),  63--86 (2013)

\bibitem{halpern2003total}
Halpern, A.C.: Total body skin imaging as an aid to melanoma detection. In:
  Seminars in cutaneous medicine and surgery. vol.~22, pp.~2--8 (2003)

\bibitem{kim2021deep}
Kim, H., Kim, J., Kam, J., Park, J., Lee, S.: Deep virtual markers for
  articulated 3d shapes. In: Proceedings of the IEEE/CVF International
  Conference on Computer Vision. pp. 11615--11625 (2021)

\bibitem{korotkov2018improved}
Korotkov, K., Quintana, J., Campos, R., Jes{\'u}s-Silva, A., Iglesias, P.,
  Puig, S., Malvehy, J., Garcia, R.: An improved skin lesion matching scheme in
  total body photography. IEEE journal of biomedical and health informatics
  \textbf{23}(2),  586--598 (2018)

\bibitem{korotkov2014new}
Korotkov, K., Quintana, J., Puig, S., Malvehy, J., Garcia, R.: A new total body
  scanning system for automatic change detection in multiple pigmented skin
  lesions. IEEE transactions on medical imaging  \textbf{34}(1),  317--338
  (2014)

\bibitem{li2016skin}
Li, Y., Esteva, A., Kuprel, B., Novoa, R., Ko, J., Thrun, S.: Skin cancer
  detection and tracking using data synthesis and deep learning. arXiv preprint
  arXiv:1612.01074  (2016)

\bibitem{mcgregor1998automatic}
Mcgregor, B.: Automatic registration of images of pigmented skin lesions.
  Pattern Recognition  \textbf{31}(6),  805--817 (1998)

\bibitem{mirzaalian2009graph}
Mirzaalian, H., Hamarneh, G., Lee, T.K.: A graph-based approach to skin mole
  matching incorporating template-normalized coordinates. In: 2009 IEEE
  Conference on Computer Vision and Pattern Recognition. pp. 2152--2159. IEEE
  (2009)

\bibitem{mirzaalian2012uncertainty}
Mirzaalian, H., Lee, T.K., Hamarneh, G.: Uncertainty-based feature learning for
  skin lesion matching using a high order mrf optimization framework. In:
  International Conference on Medical Image Computing and Computer-Assisted
  Intervention. pp. 98--105. Springer (2012)

\bibitem{mirzaalian2013spatial}
Mirzaalian, H., Lee, T.K., Hamarneh, G.: Spatial normalization of human back
  images for dermatological studies. IEEE journal of biomedical and health
  informatics  \textbf{18}(4),  1494--1501 (2013)

\bibitem{mirzaalian2016skin}
Mirzaalian, H., Lee, T.K., Hamarneh, G.: Skin lesion tracking using structured
  graphical models. Medical image analysis  \textbf{27},  84--92 (2016)

\bibitem{mitchel2021echo}
Mitchel, T.W., Rusinkiewicz, S., Chirikjian, G.S., Kazhdan, M.: Echo: Extended
  convolution histogram of orientations for local surface description. In:
  Computer Graphics Forum. vol.~40, pp. 180--194. Wiley Online Library (2021)

\bibitem{perednia1992automatic}
Perednia, D.A., White, R.G.: Automatic registration of multiple skin lesions by
  use of point pattern matching. Computerized medical imaging and graphics
  \textbf{16}(3),  205--216 (1992)

\bibitem{roning1998registration}
Roning, J., Riech, M.: Registration of nevi in successive skin images for early
  detection of melanoma. In: Proceedings. Fourteenth International Conference
  on Pattern Recognition (Cat. No. 98EX170). vol.~1, pp. 352--357. IEEE (1998)

\bibitem{saint20183dbodytex}
Saint, A., Ahmed, E., Cherenkova, K., Gusev, G., Aouada, D., Ottersten, B.,
  et~al.: 3dbodytex: Textured 3d body dataset. In: 2018 International
  Conference on 3D Vision (3DV). pp. 495--504. IEEE (2018)

\bibitem{saint2019bodyfitr}
Saint, A., Cherenkova, K., Gusev, G., Aouada, D., Ottersten, B., et~al.:
  Bodyfitr: robust automatic 3d human body fitting. In: 2019 IEEE International
  Conference on Image Processing (ICIP). pp. 484--488. IEEE (2019)

\bibitem{salti2014shot}
Salti, S., Tombari, F., Di~Stefano, L.: Shot: Unique signatures of histograms
  for surface and texture description. Computer Vision and Image Understanding
  \textbf{125},  251--264 (2014)

\bibitem{strkakowska2022skin}
Strakowska, M., Kocio{\l}ek, M.: Skin lesion matching algorithm for application
  in full body imaging systems. In: International Conference on Information
  Technologies in Biomedicine. pp. 222--233. Springer (2022)

\bibitem{strzelecki2021skin}
Strzelecki, M.H., Str{\k{a}}kowska, M., Koz{\l}owski, M., Urba{\'n}czyk, T.,
  Wielowieyska-Szybi{\'n}ska, D., Kocio{\l}ek, M.: Skin lesion detection
  algorithms in whole body images. Sensors  \textbf{21}(19), ~6639 (2021)

\bibitem{tombari2010unique}
Tombari, F., Salti, S., Di~Stefano, L.: Unique signatures of histograms for
  local surface description. In: Computer Vision--ECCV 2010: 11th European
  Conference on Computer Vision, Heraklion, Crete, Greece, September 5-11,
  2010, Proceedings, Part III 11. pp. 356--369. Springer (2010)

\bibitem{tombari2011combined}
Tombari, F., Salti, S., Di~Stefano, L.: A combined texture-shape descriptor for
  enhanced 3d feature matching. In: 2011 18th IEEE international conference on
  image processing. pp. 809--812. IEEE (2011)

\bibitem{white1992automatic}
White, R.G., Perednia, D.A.: Automatic derivation of initial match points for
  paired digital images of skin. Computerized medical imaging and graphics
  \textbf{16}(3),  217--225 (1992)

\bibitem{zhao2022skin3d}
Zhao, M., Kawahara, J., Abhishek, K., Shamanian, S., Hamarneh, G.: Skin3d:
  Detection and longitudinal tracking of pigmented skin lesions in 3d
  total-body textured meshes. Medical Image Analysis  \textbf{77},  102329
  (2022)

\end{thebibliography}

\end{document}